\documentclass[a4paper, 10pt, conference]{ieeeconf}      

\IEEEoverridecommandlockouts                              

\usepackage{graphics} 
\usepackage{epsfig} 
\usepackage{mathptmx} 
\usepackage{times} 
\usepackage{amsmath} 
\usepackage{amssymb}  
\usepackage{subcaption} 

\usepackage[backend=biber, style=ieee]{biblatex}
\title{\LARGE \bf
Prometheus: Universal, Open-Source Mocap-Based Teleoperation System with Force Feedback for Dataset Collection in Robot Learning
} 

\author{Sergei Satsevich\textsuperscript{*}, Artem Bazhenov\textsuperscript{*}, Sergei Egorov, Artem Erkhov, Maxim Gromakov, Aleksey Fedoseev, \\ and Dzmitry Tsetserukou
\thanks{\textsuperscript{*}These authors contributed equally to this work.} 
\thanks{The authors are with Intelligent Space Robotics Laboratory, CDE, Skolkovo Institute of Science and Technology, Bolshoy Boulevard 30, bld. 1, 121205, Moscow, Russia. Email: {\tt\small \{
Sergei.Satsevich, Artem.Bazhenov, Sergei.Egorov, Artem.Erhov, Maxim.Gromakov, Aleksey.Fedoseev, Dzmitry.Tsetserukou\}@skoltech.ru}}
}

\begin{document}

\maketitle
\thispagestyle{empty}
\pagestyle{empty}

\begin{abstract}
This paper presents a novel teleoperation system with force feedback, utilizing consumer-grade HTC Vive Trackers 2.0. The system integrates a custom-built controller, a UR3 robotic arm, and a Robotiq gripper equipped with custom-designed fingers to ensure uniform pressure distribution on an embedded force sensor. Real-time compression force data is transmitted to the controller, enabling operators to perceive the gripping force applied to objects. Experimental results demonstrate that the system enhances task success rates and provides a low-cost solution for large-scale imitation learning data collection without compromising affordability. 

All hardware designs, software, and implementation details—including fabrication guidelines, printed circuit board (PCB) specifications, and deployment instructions—are released as open-source at https://github.com/Eterwait/Prometheus.
\end{abstract}



\begin{figure}[htp]
    \centering
    \includegraphics[width=1\linewidth]{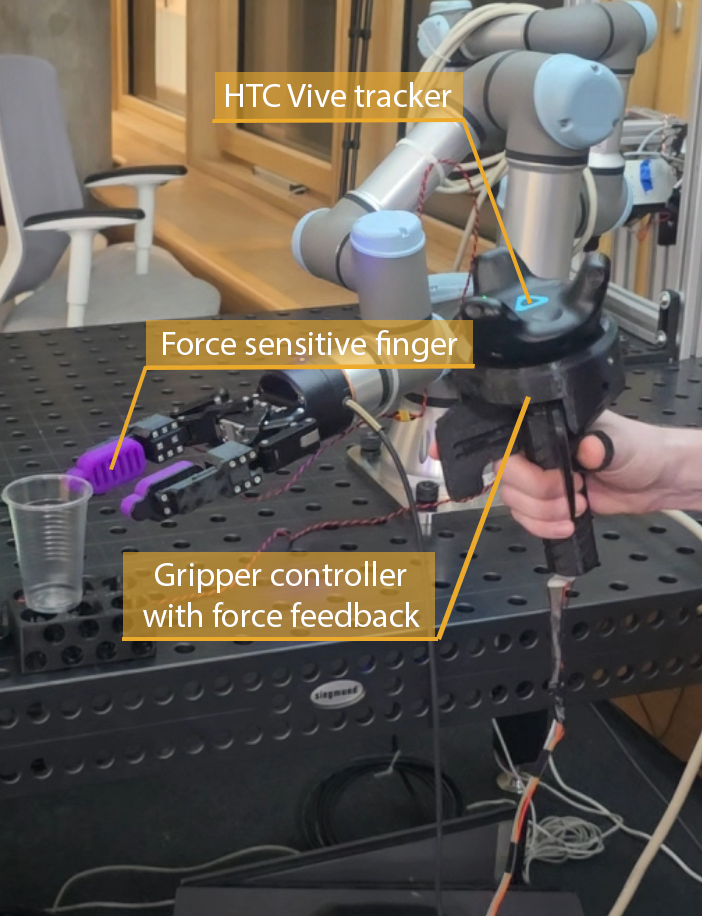}
    \caption{Main view of the system}
    \label{fig:main view}
    \vspace{0.2cm}
\end{figure}

\section{Introduction}
Robotics has undergone significant advancements in recent years, particularly in demonstration-based learning using neural networks with transformer architectures. These methods show remarkable potential for automating both industrial and everyday tasks. However, the performance of such models heavily depends on the quantity and quality of the training data.

A common data collection approach involves teleoperation with motion capture systems, where human motions are transferred to a robot via inverse kinematics. While effective, this method has a critical limitation: the lack of force feedback during object manipulation. Without accurate force perception, both the operator and the neural network may apply excessive gripping forces, leading to object deformation or damage.

To address this issue, we propose a low-cost, open-source teleoperation system with integrated force feedback (Fig. 1). Our solution leverages 3D-printed and commercially available mechanical components, combined with custom-designed PCBs. \textbf{The entire system—including all hardware, software, and firmware components— will be made available as open-source resources, including detailed instructions for fabrication, assembly, PCB ordering, and system deployment.}

\section{Related Work}

\subsection{Teleoperation}

Modern approaches to teleoperation can generally be divided into two categories.

The first approach is robot-specific and relies on building a kinematic copy of the robot. By knowing the robot’s joint positions, the system uses forward kinematics to replicate the robot’s behavior. This method provides a natural, intuitive way of controlling the robot by directly mirroring its structure and movement. Examples of this approach can be seen in systems like Echo \cite{b1}, GELLO \cite{b2}, ALOHA \cite{b3}, Mobile ALOHA \cite{b4}, AirExo \cite{b5}, and Aloha 2 \cite{b6}.

The main advantage of this approach is its intuitiveness, as operators can directly interact with a system that mirrors the robot’s structure, requiring minimal learning. Furthermore, it offers high precision in tasks demanding fine motor skills, such as surgical operations or delicate industrial processes. However, the robot-specific nature of this approach limits its adaptability; each system must be custom-built for the robot it controls. Additionally, the need for a physical kinematic copy increases costs and technical complexity, making this approach less practical for more generalized applications.

The second approach is more versatile and can be used across a variety of robotic systems. It employs inverse kinematics to reproduce the end-effector’s position, bypassing the need for a direct kinematic copy. The position of the end-effector can be determined using VR controllers \cite{b7, b8, b9, b10, b11, b12, b13, b14, b15, b16, b17, b19}, glove-based systems \cite{b18, b19, b20, b21, b22, b23}, vision-based tracking \cite{b22, b24, b25, b26, b27, b28, b29, b30}, or IMU sensors \cite{b6, b37}. This method is particularly useful in scenarios where greater flexibility and system compatibility are required.

Unlike the first approach, this method shines in its adaptability and applicability to a wide range of robots. Input devices like VR controllers or gloves provide flexibility, enabling operators to control different robotic systems without the need for custom hardware. However, it comes with certain drawbacks. The reliance on inverse kinematics introduces computational complexity, which may lead to delays or inaccuracies in real-time scenarios. Additionally, while versatile, this approach can feel less intuitive to operators and may require training to achieve precise control.

\subsection{Haptic feedback}

Haptic feedback systems enhance teleoperation by providing force and tactile sensations to the operator, improving precision and immersion. Several approaches exist for integrating haptic feedback with bimanual manipulators, utilizing different control devices, each with its own set of limitations.

One notable approach uses controllers equipped with repurposed prosthetic hands, as demonstrated in \cite{b7}. These systems leverage touch sensors embedded in prosthetic hands to provide tactile feedback. While effective for simulating touch, this method is often constrained by the limited range of force sensations it can replicate, which may reduce the realism of interactions.

Another common implementation involves VR controllers, as seen in \cite{b15}, where the vibration mechanism in controllers serves as haptic feedback. This approach is cost-effective and widely accessible but offers only basic feedback through vibrations, lacking the nuanced force sensations needed for more complex tasks.

Glove-based systems, such as those used in \cite{b23, b38, b39} represent a more immersive option. These systems provide a combination of tactile and force feedback by mimicking the hand’s natural movements. However, gloves often suffer from limited precision and can be cumbersome for extended use, particularly in tasks requiring fine motor skills.

Specialized haptic devices, highlighted in works like \cite{b40, b41}, offer the most precise and responsive feedback. These devices excel in delivering a high-fidelity force-feedback experience, making them ideal for critical applications such as medical simulations. However, their high cost and complexity limit their adoption in broader applications.

\section{System Architecture}

Our system consists of a custom-built controller with HTC VIVE Tracker 2.0, a UR3 robot, and a Robotiq 2F-85 gripper equipped with custom robotic fingers for uniform pressure distribution on the internal force sensor. The operator uses HD Pro Webcam C920 USB, mounted on the gripper for visual feedback, and the controller for force feedback.  

\subsection{Controller}

\begin{figure}[t]
    \centering
    \includegraphics[width=0.6\linewidth]{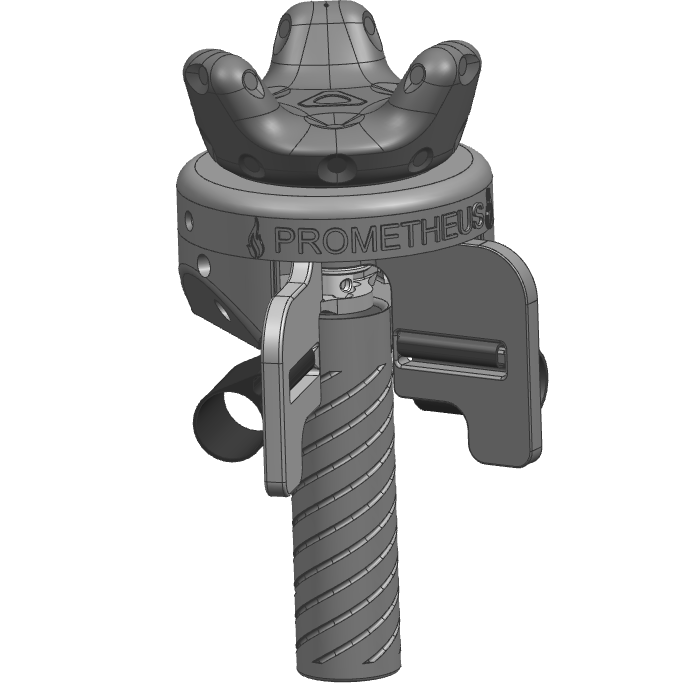}
    \caption{External view of the controller}
    \label{fig:controller}
    \vspace{0.1cm}
\end{figure}

To receive force feedback and ensure comfortable control of the robot and gripper, we developed a specialized haptic device (Fig.~\ref{fig:controller}). It consists of a frame connected to a handle and a cover with an attached HTC Vive tracker. Two actuation sticks rotate on bearings mounted on the frame. One stick is connected to the motor's frame, while the other is connected to the motor's shaft.

\begin{figure}[t]
    \centering
    \includegraphics[width=0.6\linewidth]{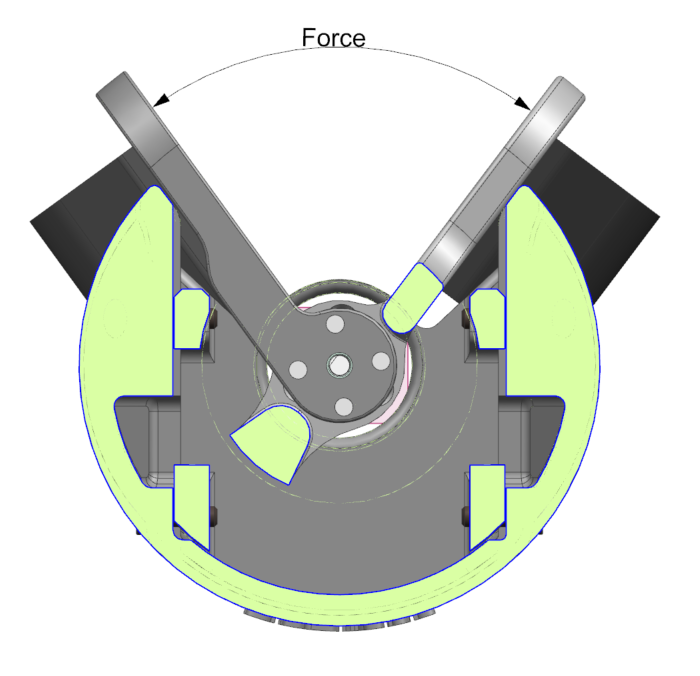}
    \caption{Principle of work}
    \label{fig:principle}
    \vspace{0.2cm}
\end{figure}

The working principle of this controller is as follows (Fig.~\ref{fig:principle}): when the motor is mounted on a bearing and a lever is attached to its shaft, this produces torque on the motor shaft. According to Newton's Third Law, the motor's body experiences an equal and opposite reaction torque. Since the motor is not fixed to an external support, it begins to rotate in the opposite direction. As a result, the lever and the motor body rotate in opposite directions.

\begin{figure}[t]
    \centering
    \includegraphics[width=0.7\linewidth]{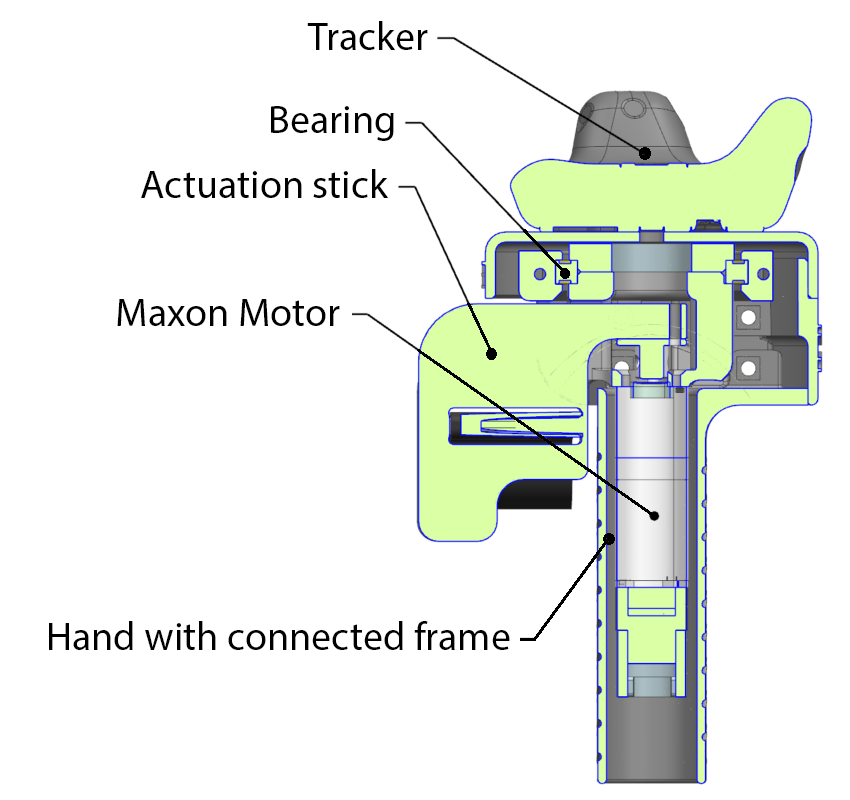}
    \caption{Internal schematic of the controller}
    \label{fig:controller_int}
    \vspace{0.2cm}
\end{figure}

The controller serves as a mounting point for the HTC Tracker 2.0, which is used to track the coordinates of the human hand. Using the approach described in \cite{b8}, we transform the position and orientation of the hand from the operator's frame to that of the manipulator. The Real-Time Data Exchange library \cite{b36} is used to calculate the inverse kinematics of the UR3 and transmit the corresponding joint angles to the manipulator.

\subsection{Force sensitive finger}

\begin{figure}[t]
    \centering
    \includegraphics[width=0.6\linewidth]{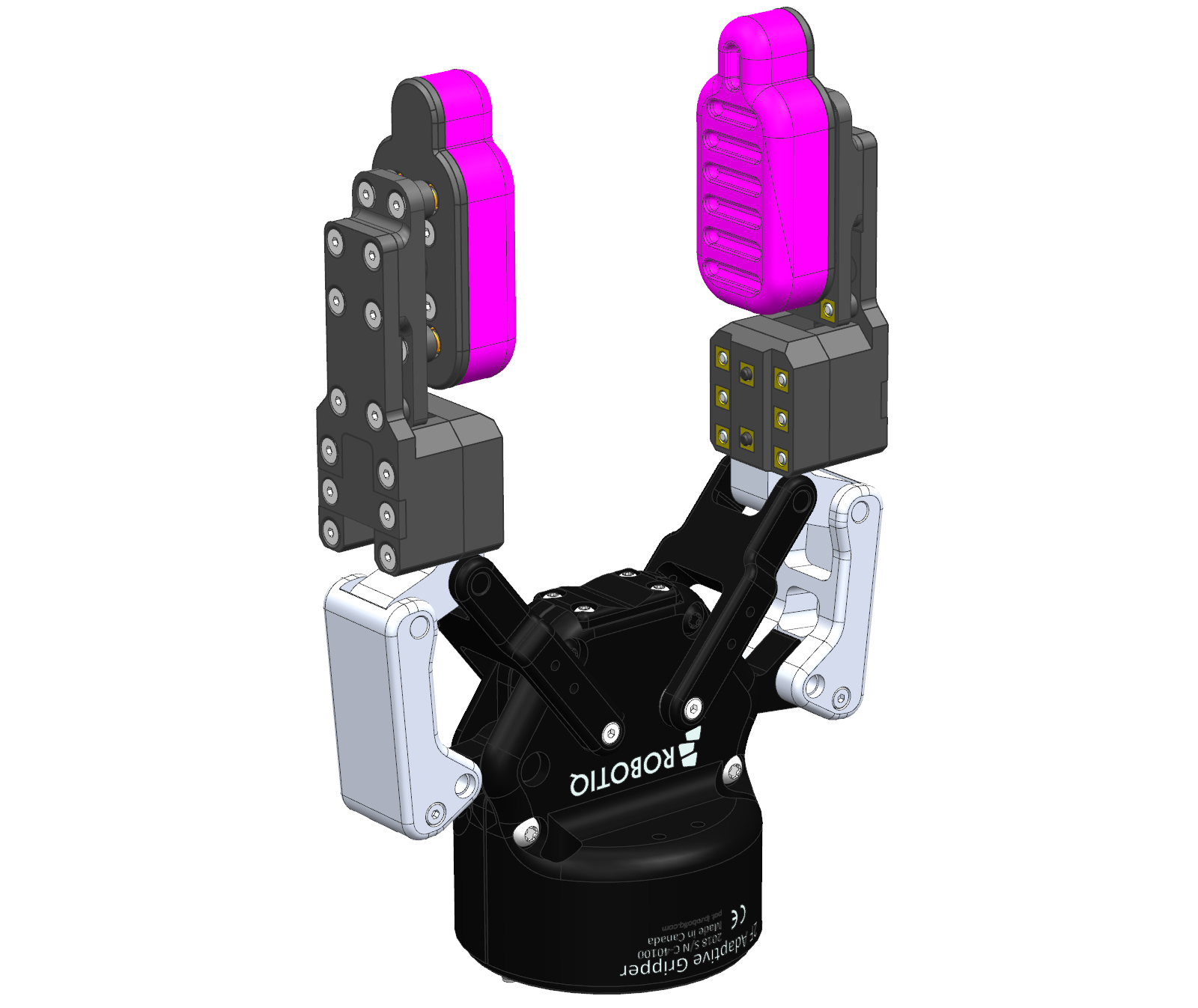}
    \caption{Gripper with a force-feedback robotic finger}
    \label{fig:gripper}
    \vspace{-0.4cm}
\end{figure}

For gripping operations and receiving force feedback, we placed a force sensor in the gripper finger (Fig.~\ref{fig:gripper}). We used the Robotiq 2F-85 as the gripper and the RP-C7.6-LT as the force sensor. The main challenge in integrating a force sensor into robotic gripper end effectors lies in determining an optimal placement. Directly gluing the force sensor onto the gripper’s end \cite{b31} is not an ideal solution, as it limits force feedback to a small sensing area and increases the risk of sensor damage.

\begin{figure}[t]
    \centering
    \includegraphics[width=0.7\linewidth]{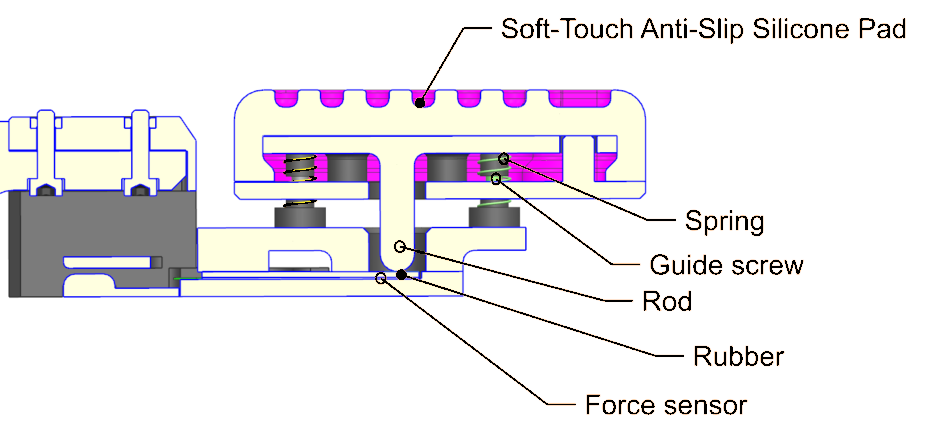}
    \caption{Schematic of the robotic finger mechanism}
    \label{fig:finger_mech}
    \vspace{0.2cm}
\end{figure}

One possible solution is the development of a custom sensor \cite{b32} based on a spring and potentiometer integrated into the gripper’s structure. In our work, we designed a unique gripper finger that provides force feedback from the entire surface of the end effector while enhancing adaptability and protecting the force sensor (Fig.~\ref{fig:finger_mech}).

\begin{figure}[h]
    \centering
    \begin{subfigure}[b]{0.6\linewidth}
        \centering
        \includegraphics[width=\linewidth]{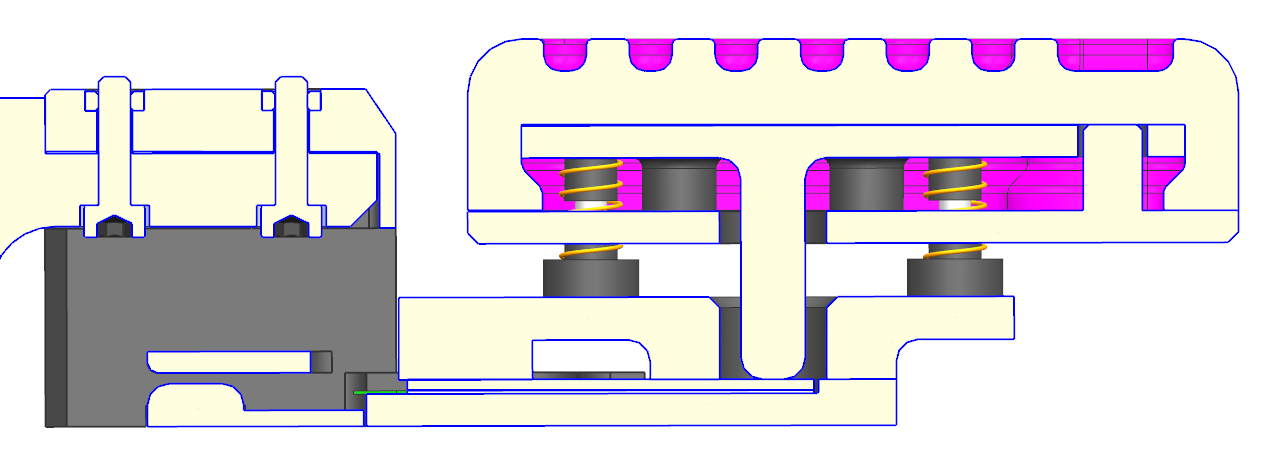}
        \caption{Initial position}
        \label{fig:sub1}
    \end{subfigure}
    
    \begin{subfigure}[b]{0.6\linewidth}
        \centering
        \includegraphics[width=\linewidth]{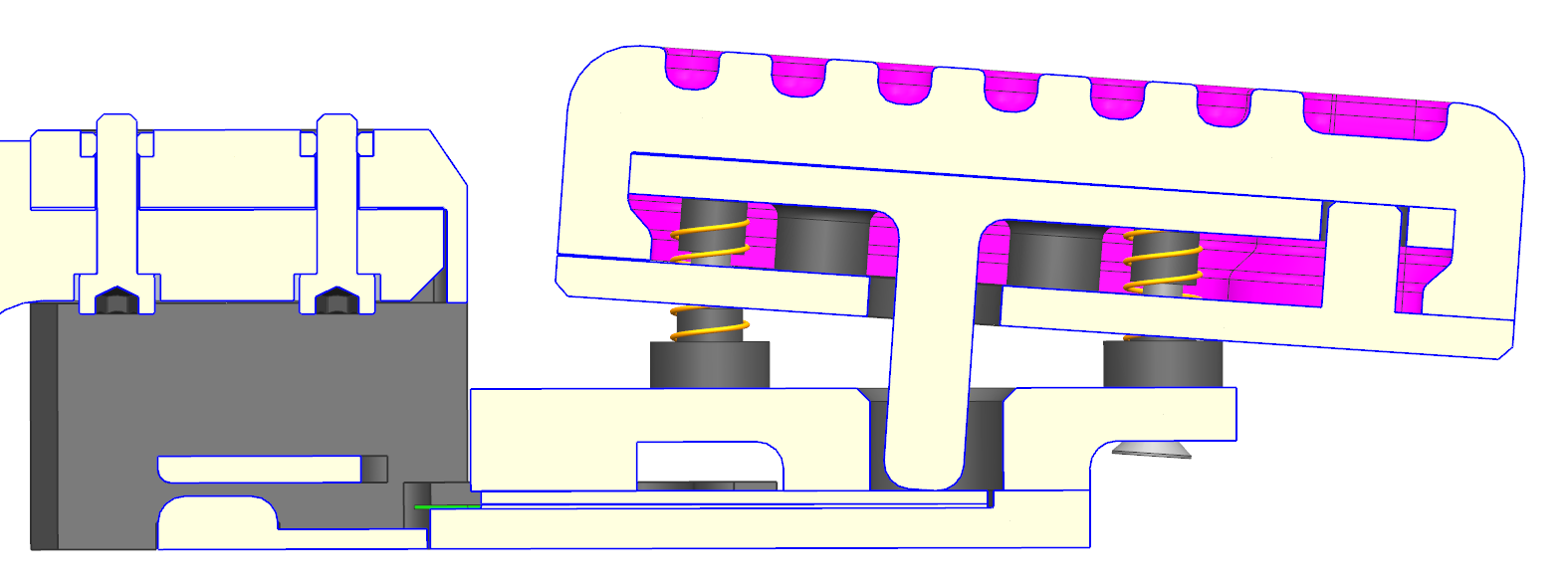}
        \caption{Force applied to the end of the pad}
        \label{fig:sub2}
    \end{subfigure}

    \begin{subfigure}[b]{0.6\linewidth}
        \centering
        \includegraphics[width=\linewidth]{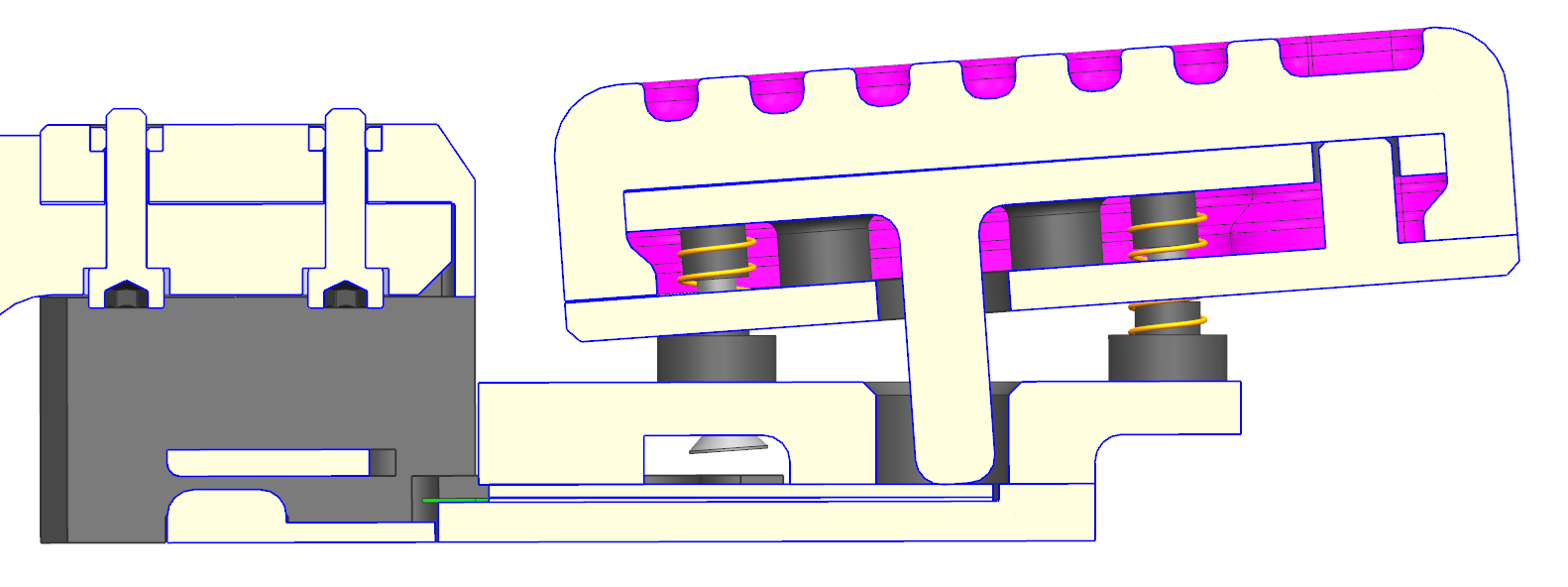}
        \caption{Force applied to the beginning of the pad}
        \label{fig:sub2}
    \end{subfigure}

    \caption{Operation of the mechanism under applied force}
    \label{fig:mechanism}
\end{figure}

The working principle of our mechanism involves pushing a rod, which is part of the pad structure, onto the force sensor through a rubber sheet (Fig.~\ref{fig:mechanism}). Springs return the pad to its original position after the force is removed, while guide screws secure the pad to the structure. The Soft-Touch Anti-Slip Silicone Pad was cast from two-component silicone into a plastic mold. This pad prevents slipping, ensures more uniform force distribution, and enables soft gripping of objects.

\subsection{Electronics}
The system consists of PCBs: the Force Sensor Board and the Motor Control Board. The Force Sensor Board measures the force exerted by the gripper, linearizes the signal, and transmits it to the Motor Control Board. The Motor Control Board then computes the control force required for the motor, processes encoder position data, and transmits all relevant information to a computer.
\begin{figure}[h]
    \centering
    \begin{subfigure}[b]{1\linewidth}
        \centering
        \includegraphics[width=\linewidth]{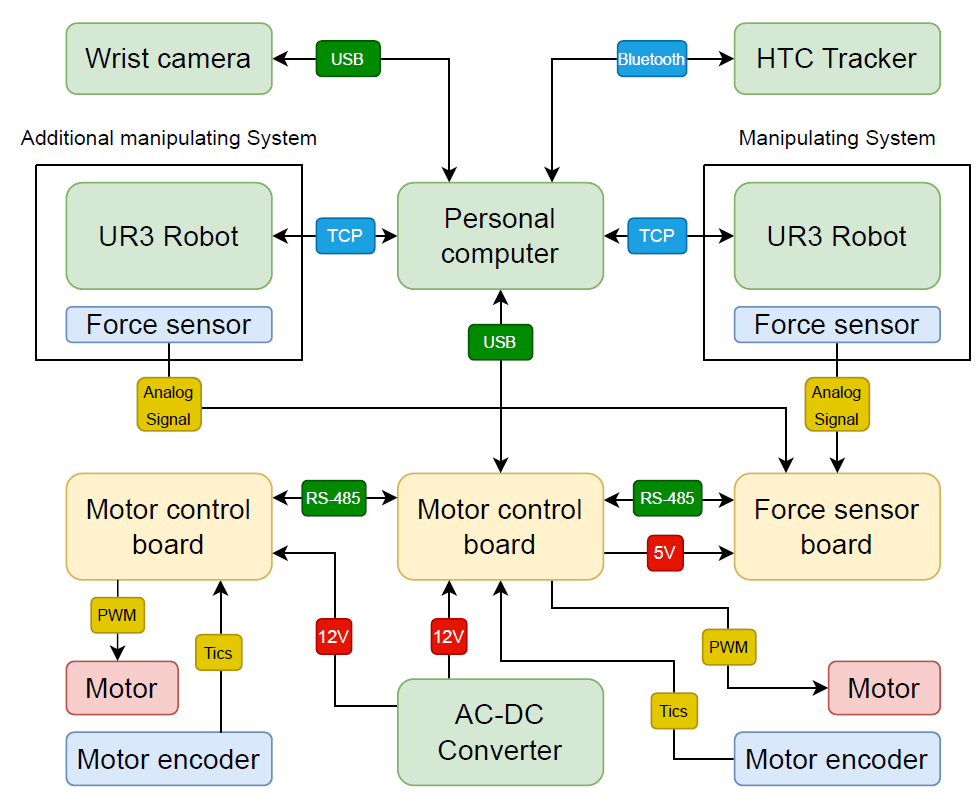}
        \label{fig:sub2}
    \end{subfigure}
    \caption{System architecture}
    \label{fig:architecture}
\end{figure}
The Motor Control Board is based on an STM32F401RET6TR MCU, which manages all system operations. The board is equipped with reverse polarity and overcurrent protection. It communicates with the computer via a galvanically isolated USB interface and exchanges data with the Force Sensor Board using the RS-485 protocol.

The Force Sensor Board is based on an STM32F303CBT6 MCU. It receives power and control commands from the Motor Control Board, acquires data from the force sensor, linearizes the signal using an operational amplifier with a bipolar power supply, and transmits the processed data back to the Motor Control Board.



\subsubsection{Force Sensor Linearization System}
The employed force sensors operate based on the principle of decreasing electrical resistance under applied pressure. However, the resulting voltage response is nonlinear and requires linearization. This is achieved using an operational amplifier with a bipolar power supply. The force sensor serves as the input to a current-to-voltage converter, whose output is governed by the following equation:
\begin{equation}
V_{\text{OUT}} = V_{\text{REF}} \times \left( -\frac{R_G}{R_{\text{FS}}} \right)
\end{equation}

where:  
\begin{itemize}
    \item \( V_{\text{REF}} = 3.3 \) V is the reference voltage applied to the operational amplifier.
    \item \( R_G \) is the feedback resistor in the operational amplifier, which determines the gain of the circuit.
    \item \( R_{\text{FS}} \) is the resistance of the force sensor in its fully compressed state.
\end{itemize}

Using this equation, you can select resistors for force sensors with different force feedback characteristics.

\section{User Study}

\subsection{Experimental Setup}

\begin{figure}[h]
    \centering
    \begin{subfigure}[b]{0.7\linewidth}
        \centering
        \includegraphics[width=\linewidth]{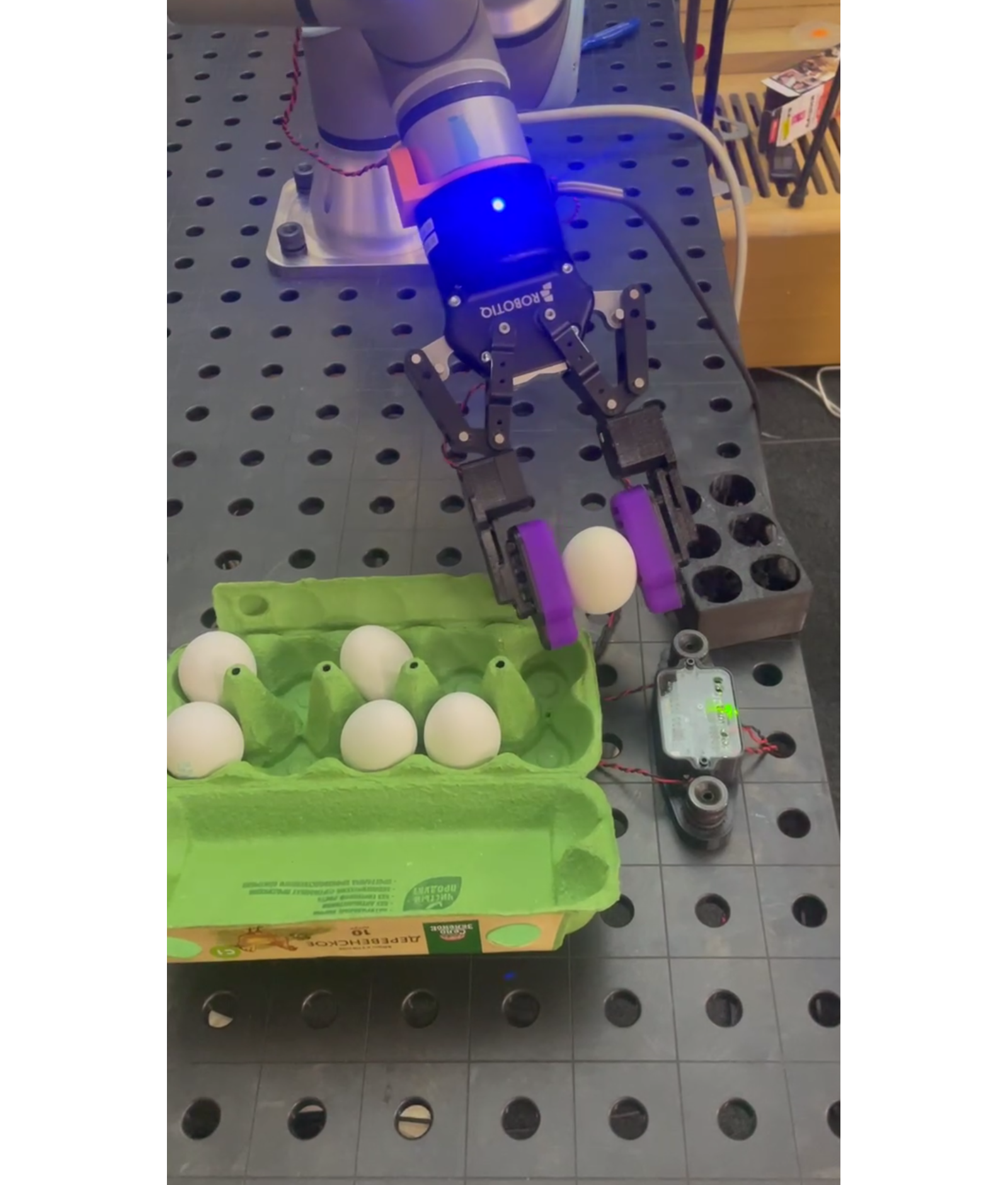}
        \label{fig:exp}
    \end{subfigure}
    \caption{Eggs experiment}
    \label{fig:eggs}
\end{figure}

A series of experiments were carried out to compare the amount of force necessary for pick-and-place tasks while using the proposed system Prometheus (Fig.~\ref{fig:eggs}). Users were asked to perform a pick-and-place task with an egg in two scenarios: (1) with force feedback and (2) without force feedback.

Participants: Fifteen participants, six females and nine males, capable of performing the experiment, aged 24.1±4.3 years, volunteered to participate in the experiments.

\subsection{Experimental Results}

\begin{figure}[h]
    \centering
    \begin{subfigure}[b]{0.8\linewidth}
        \centering
        \includegraphics[width=\linewidth]{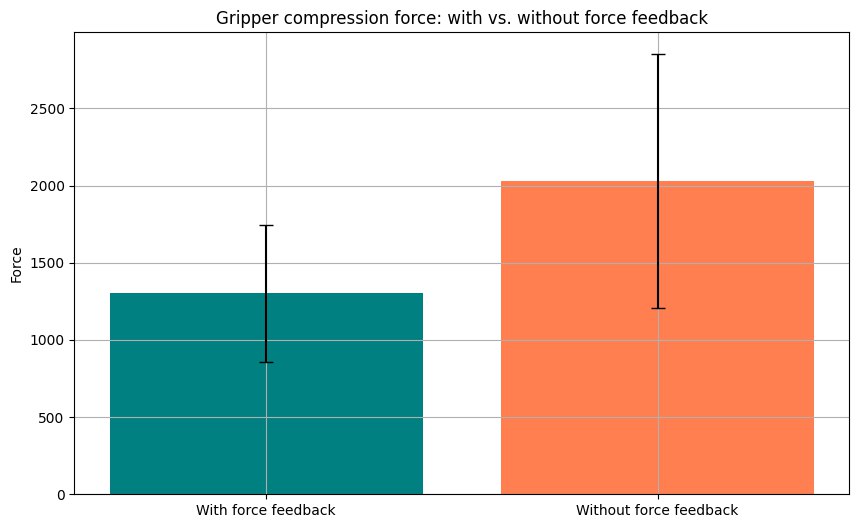}
        \label{fig:exp}
    \end{subfigure}
    \caption{Gripper compression force: with vs. without force feedback}
    \label{fig:graph}
\end{figure}

The results indicate that, on average, users applied 35.77\% less force in the first scenario compared to the second when grasping and manipulating the object (Fig.~\ref{fig:graph}). This finding demonstrates that force feedback enables users to apply only the necessary amount of force for object handling, which is particularly important when working with fragile items.

\section{Compatibility with VLA models}

This section evaluates the impact of force feedback data — collected using our teleoperation system alongside standard robot proprioceptive states and actions — on the performance of fine-tuned Vision-Language-Action (VLA) models. To isolate the contribution of force sensing, we trained three variants of the same base model, each with distinct proprioceptive observation spaces:
\begin{itemize}
    \item Position-only (P): Standard image and task instruction inputs with continuous gripper position control (no force feedback).
    \item Force-only (F): Force sensor input added, but gripper state information excluded.
    \item Position-and-force (P+F): Integrated gripper position and force feedback.
\end{itemize}

\subsection{Dataset structure}

We collected a 300-trajectory dataset across three tasks requiring precise grasp force modulation. Each task was designed to test the model’s ability to handle objects with varying deformability. 

For all tasks, state observation consists of:
\begin{itemize}
    \item An RGB image (128 × 128) captured from a wrist-mounted camera,  
    \item An RGB-D image (256 × 256) from a RealSense D455 side camera,  
    \item The robot’s proprioceptive state, which includes end-effector pose and joint angles, \item Gripper position and force sensor readings, standardized to the range [0, 1].
\end{itemize}

The action space is defined as a 7-dimensional vector, comprising:
\begin{itemize}
    \item A 6-dimensional delta for joint angles,
    \item A 1-dimensional delta for gripper position, standardized to the range [-1, 1]. 
\end{itemize}

The frequency of data collection was fixed at 10Hz.

\subsection{Tasks}

The selected objects for manipulation included a shampoo bottle, a tomato, and a tube of toothpaste. Each trajectory starts from the robot's base position and ends when the robot successfully grasps the object and lifts it approximately 10 cm above the table.

The robot initiated each trial from a fixed base position, with targets placed within a 30 cm semicircle beneath the gripper. Objects were distributed uniformly across this workspace. We maintained consistent background conditions with controlled artificial lighting. Orientation protocols varied: shampoo (±15° from default), toothpaste (random orientations converging to two stable positions during grasping), and tomatoes (random orientation with ±5 mm size variation).

\subsection{Model Selection and Training}

We use Octo-Small \cite{b43}, a pre-trained transformer-based model, which provides a balance between performance and inference efficiency while maintaining a compact architecture.

The proprioceptive inputs were uniformly discretized into 256 bins and tokenized using Octo's LowDim tokenizer. Following \cite{b44}, we adopted a delta joint angles action space, a linear action head, and a full fine-tuning strategy.

Each policy was exclusively trained on task-specific trajectories (100 episodes per object). We employed 50,000 training epochs for model optimization, though empirical observations revealed that performance typically plateaued after 20,000 iterations, with no significant improvements thereafter.

The complete fine-tuning process required 2.5 hours to complete 50,000 epochs using a batch size of 32 on an NVIDIA RTX 4090 GPU with 24GB of VRAM.

\subsection{Experimental results}

\begin{figure}[h]
    \centering
    \includegraphics[width=1.0\linewidth]{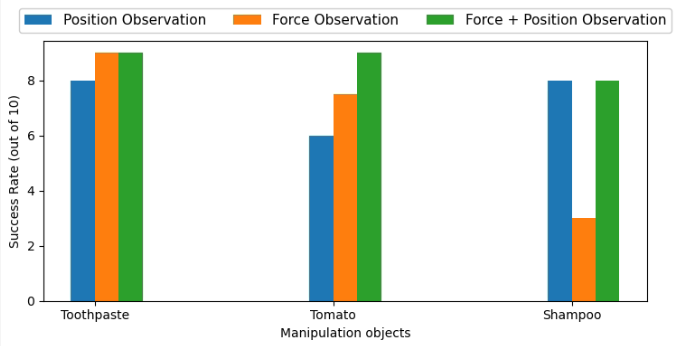}
    \caption{Success rates across three policy variants (position-only, force-only, and position-and-force) evaluated on three object types: toothpaste tube, tomato, and shampoo bottle. Each policy was tested in 10 independent trials per object.}
    \label{fig:success_rates}
    \vspace{0.2cm}
\end{figure}

Results were averaged among 10 trials per object-policy combination (Fig.~\ref{fig:success_rates}). 

\begin{figure}[h]
    \centering
    \includegraphics[width=1.0\linewidth]{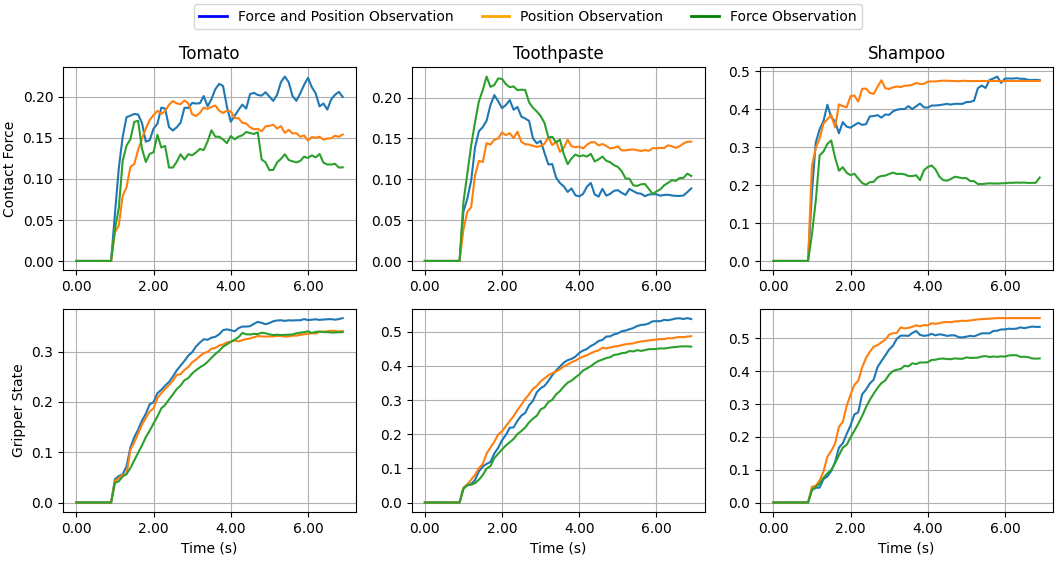}
    \caption{Time-series analysis of grasp execution dynamics across three manipulation tasks. Each task is represented by two subplots: (a) continuous gripper position and (b) measured contact force. Both are standardized to the range [0, 1]. Curves show mean values averaged over 10 trials for three policy variants: position-only (orange), force-only (green), and position-and-force (blue).}
    \label{fig:plots}
    \vspace{0.2cm}
\end{figure}

\textit{Tomato}. Both force-only and position-only policies exhibited slippage in 20\% of trials. The position-only variant additionally caused irreversible deformation in 20\% of cases by applying excessive force (3× the average required). In contrast, the P+F policy maintained stable grasps without damage (90 \%).

\textit{Toothpaste}. All three policy variants achieved comparable success rates (~80–90\%), demonstrating minimal performance divergence. This suggests that the tube’s elastic properties mitigate the consequences of force misapplication: even excessive grip forces (observed in position-only trials) did not cause damage, while insufficient forces (typical of force-only and force + position trials) still allowed successful lifts due to the object’s light weight and deformable structure. Unlike rigid (shampoo bottle) or highly deformable (tomato) objects, the toothpaste tube’s intermediate compliance rendered force feedback less critical for task success, though it remained essential for precise force regulation in edge cases (e.g., near-empty tubes).

\textit{Shampoo bottle}. The force-only policy unexpectedly underperformed (30\% success), consistently applying insufficient force and leading to slippage (Fig.~\ref{fig:plots}). Both position-only and P+F policies achieved 90\% success, suggesting that gripper position is the dominant signal for rigid or minimally compressible objects. 

\section{Discussion}
Our experimental results highlight the benefits of incorporating force feedback in teleoperation systems for imitation learning data collection. In the user study, participants applied 35.77\% less gripping force on average when force feedback was enabled, demonstrating improved precision in handling fragile objects like eggs. This reduction aligns with prior research on haptic interfaces \cite{b40, b41}, which shows that tactile cues enhance operator awareness and reduce errors in manipulation tasks. Compared to vibration-based feedback in VR controllers \cite{b15}, our torque-based mechanism provides more nuanced force perception, potentially leading to safer interactions in real-world scenarios.

The integration with Vision-Language-Action (VLA) models further underscores the value of force data. Policies trained with both position and force inputs (P+F) achieved up to 90\% success rates across tasks involving deformable objects, outperforming position-only (P) and force-only (F) variants in cases like tomato grasping, where excessive force caused deformation or slippage. These findings extend related work on multi-modal robot learning \cite{b43, b44}, suggesting that force sensing complements visual and proprioceptive data, enabling models to learn more adaptive grasping behaviors. However, the force-only policy's underperformance on rigid objects like shampoo bottles indicates that force feedback may be redundant in low-deformability scenarios, emphasizing the need for task-specific sensor fusion strategies.

Despite these advancements, our system has notable limitations. It is currently optimized for the UR3 arm and Robotiq 2F-85 gripper, limiting direct applicability to other robotic platforms without modifications. The dataset for VLA fine-tuning was modest (300 trajectories across three tasks), which may constrain generalization to unseen objects or environments. Additionally, while latency in inverse kinematics and sensor processing was not measured, it could impact real-time performance in dynamic settings. Cost-effectiveness relies on consumer-grade components, but custom PCB fabrication may pose barriers for non-expert users, despite open-source resources.

The implications of this work extend to broader applications in embodied AI, where high-quality, force-aware datasets can improve the robustness of foundation models for household and industrial robotics. By addressing the lack of haptic feedback in motion capture-based teleoperation \cite{b7, b8}, Prometheus facilitates scalable data collection without specialized hardware.

Future work could involve porting the system to diverse robots, such as dual-arm setups for bimanual tasks, and incorporating additional modalities like tactile textures or vibration for richer feedback. Larger-scale datasets and evaluations on long-term operator fatigue would further validate its utility. Exploring hybrid learning approaches, combining imitation with reinforcement learning, could leverage force data to accelerate policy optimization.

\section{Conclusion}

In this paper we present Prometheus, an open-source, low-cost teleoperation system integrating force feedback to enhance data collection for imitation learning. By combining HTC Vive Trackers, a custom haptic controller, and a modified Robotiq gripper, the system enables precise manipulation with real-time force perception, reducing applied gripping forces by 35.77\% and improving success rates in Vision-Language-Action model training up to 90\% for fragile objects. The incorporation of force feedback proves particularly valuable for handling deformable objects, preventing damage and slippage, and enriching datasets for robust robot learning. With fully open-source hardware and software, Prometheus supports community-driven advancements, offering a scalable solution for improving human-robot interaction in both industrial and domestic applications.

\printbibliography

\end{document}